\documentclass[fleqn,10pt]{wlscirep}
\usepackage[utf8]{inputenc}
\usepackage[T1]{fontenc}
\usepackage{mathtools}
\usepackage{subcaption}
\usepackage{tabularx}
\usepackage{multirow}
\usepackage{xcolor}
\usepackage{amsmath}

\title{Federated Fairness Analytics: Quantifying Fairness in Federated Learning}

\author[1,*]{Oscar Dilley}
\author[1]{Juan Marcelo Parra-Ullauri}
\author[1]{Rasheed Hussain}
\author[1]{Dimitra Simeonidou}
\affil[1]{Smart Internet Lab, School of Electrical, Electronic and Mechanical Engineering, University of Bristol, UK, BS8 1UB}
\affil[*]{oscar.dilley@bristol.ac.uk}

\begin{abstract}

Federated Learning (FL) is a privacy-enhancing technology for distributed ML. By training models locally and aggregating updates - a federation learns together, while bypassing centralised data collection. FL is increasingly popular in healthcare, finance and personal computing. However, it inherits fairness challenges from classical ML and introduces new ones, resulting from differences in data quality, client participation, communication constraints, aggregation methods and underlying hardware. Fairness remains an unresolved issue in FL and the community has identified an absence of succinct definitions and metrics to quantify fairness; to address this, we propose \emph{Federated Fairness Analytics} - a methodology for measuring fairness. Our definition of fairness comprises four notions with novel, corresponding metrics. They are symptomatically defined and leverage techniques originating from XAI, cooperative game-theory and networking engineering. We tested a range of experimental settings, varying the FL approach, ML task and data settings. The results show that statistical heterogeneity and client participation affect fairness and fairness conscious approaches such as Ditto and q-FedAvg marginally improve fairness-performance trade-offs. Using our techniques, FL practitioners can uncover previously unobtainable insights into their system's fairness, at differing levels of granularity in order to address fairness challenges in FL. We have open-sourced our work at: \url{https://github.com/oscardilley/federated-fairness}. \\

\textbf{\textit{Keywords -}} Federated Learning, Fairness, XAI, Federated Analytics, Distributed Machine Learning

\end{abstract}
\begin{document}
\flushbottom
\maketitle
\thispagestyle{empty}
\section*{Introduction}

Data is naturally distributed around us, commonly made inert by data protection regulations. Classically, to train machine learning models, data is collected centrally – the data is brought to the model. However, such scenarios where data and compute power can be centralised are not archetypal of most of the world around us – where data and processing infrastructure naturally exist in distributed pockets. This segmentation of data can be observed in many areas of modern life and it is common that the data is immovable between segments due to legislative rules. For example, corporations with data centres spread globally may be prevented from cross-border sharing due to regulation. Alternatively, in the UK’s NHS, which consists of thousands of individual practices, patient information is stored on different systems which are not typically interconnected\cite{NHScare2023}. Segmentation also arises at in personal computing where users are not willing to share their personal data with third parties. These examples only scratch the surface of the possible distributed data scenarios and motivates the idea that many parties could benefit from the application of machine learning and privacy-preserving knowledge-sharing across the different segments, instead of taking the data to the model. \\

Federated Learning was proposed in the seminal paper by B. McMahan from Google in 2017\cite{McMahan2017}. It presents the original approach for FL, FedAvg, and offers a privacy-preserving solution to the problem above, whilst being competitive in performance with its centralised counterparts in a set of typical machine learning tasks. Since its inception, FL has been deployed into a wide range of applications with cross-device implementations such as with Google’s GBoard \cite{Hard2018}; it is also rumoured, to be used in detecting the wake-up phrase, ‘Hey Siri’ for Apple’s voice agent\cite{Hao2019}. The effectiveness of cross-silo scenarios have also been demonstrated, such as in a case where a small number of UK hospitals collected data and trained a model using FL to screen for Covid-19\cite{Soltan2023}. \\

In recent, general-FL literature, fairness is repeatedly recognised as one of the key open problems. A 2021 review by P. Kairouz and B. McMahan of Google, designates a section to identifying fairness as a key area for future research in FL\cite{Kairouz2021}. Other general surveys, from 2022 and 2023, also draw attention to the potential for heterogeneity to lead to unfairness, motivating further research\cite{Ma2022, Mayhoub2023}. Finally, in a talk at the 2024 Flower AI Summit, the Head of Samsung AI Europe, refers to the difficulty managing bias and fairness in FL as “a completely unsolved problem, even in central learning”\cite{Hospedales2024}. For commercial viability of machine learning solutions, compliance with local data protection regulations such as GDPR is essential. As fairness is a core principle of GDPR\cite{Truong2021}, regulation abidance and fairness handling are inherently intertwined. This is threatened by the 'black-box' nature of many machine learning systems, where outputs are rarely transparent or justified - addressing fairness and explainability are therefore crucial for delivering industry-ready FL solutions. Fairness concerns can arise in scenarios where there are differences between client’s, behaviour, treatment, data or any other characteristic in a system. A thorough definition of fairness in FL is provided later - for now, it can be considered philosophically as equal treatment for all involved. Difference is natural in real-world systems and its effects must be anticipated to appear in many ways; we adopted the definitions of Ye et al. which suggest four flavours of heterogeneity that can impact a system - statistical, model, communication or device centric\cite{Ye2023}. \\

Several fairness-specific surveys, published in 2023, offer a general picture of the latest work in the field of FL fairness and evidence the demand for research into how such fairness can be effectively defined and quantified. Each motivates the requirements for fairness, proposes its own set of definitions and discusses the existing frameworks which have been proposed to improve fairness\cite{Vucinich2023, Shi2023, Rafi2023}. Evidencing scenarios that would benefit from fairer FL frameworks is a strength of the existing literature. For example, FL could be used to process images with the aim of detecting skin cancer, via training a classifier model over a set of globally distributed clients. Clients could range from individuals with single-image datasets, to large hospitals. It is suggested that unfairness may arise if clients achieve different levels of classification accuracy, if the model behaves favourably or unfavourably towards samples with specific characteristics – such as skin colour - or if the system exhibits bias towards certain client's hardware or network capabilities\cite{Vucinich2023}. Shi et al. go further, suggesting that it is important for the contribution of the client to be considered in attribution of fairness\cite{Shi2023}. The free-rider problem (also discussed by Rafi et al.\cite{Rafi2023}) can emerge if a client does not usefully contribute to the training of the model, but still benefits from the work of other clients when it receives updated global models. This is not a complete description of all the flavours of unfairness that can arise but it demonstrates that fairness in FL is a multi-faceted problem and that mitigation is crucial for real-world systems. It should also be noted, that there is typically a trade-off that must be considered between communication efficiency, privacy, accuracy and fairness\cite{Rafi2023, Kairouz2021}. \\

This work highlights the limitations of the current literature in defining fairness in FL and hence proposes Federated Fairness Analytics, a methodology for quantifying fairness in FL systems. With our modular notions, the fairness performance of existing systems can be evaluated in a repeatable, human-interpretable way – enabling transparency, comparison of different approaches and directed development to improve systems. Quantifying and explaining unfairness is a key, first step towards ensuring future deployments are fair and towards explainable-artificial-intelligence (XAI) in FL. The aim in this work is not to improve fairness, but to offer a platform to interpret and identify unfairness, this entailed:
\begin{itemize}
    \item Conceiving succinct, logical definitions for fairness, which are applicable to most typical FL systems.
    \item Designing corresponding normalised metrics for quantifying fairness.
    \item Developing a framework and pipeline to measure fairness during training and visualise results.
    \item Benchmarking and evaluting the fairness performance of Twenty-Four different systems - comparing the effects of differing models and datasets, heterogeneity and FL approaches.
\end{itemize}

\section*{Technical Background}

\begin{figure}[h!] %options are : !htbp
    \centering
    \includegraphics[scale=0.42]{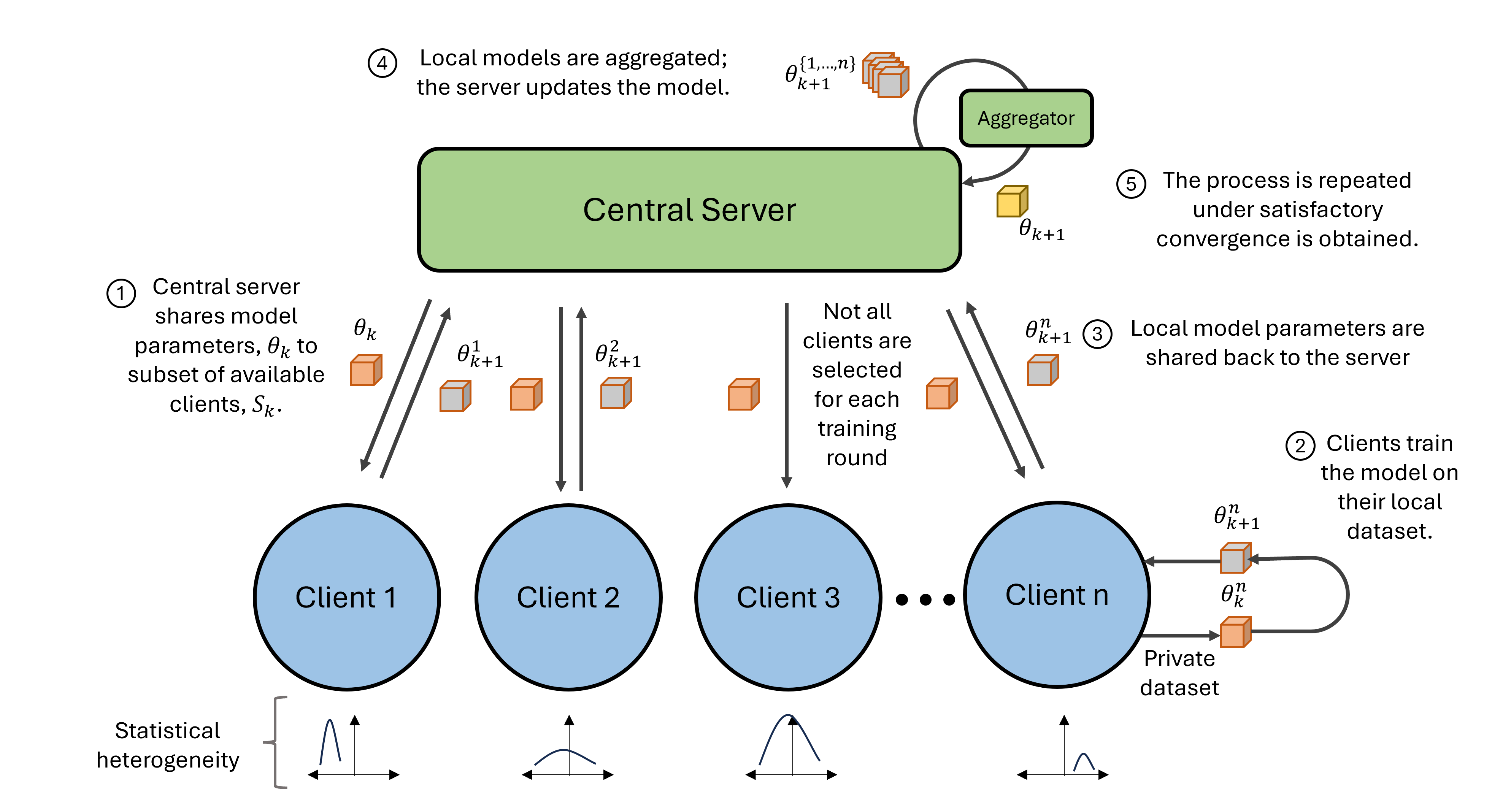}
    \caption{\centering The client-server architecture of a typical FL system with a central server and clients that exhibit statistical heterogeneity due to differing datasets.}
    \label{fig:generalFL}
\end{figure}

To form a technical definition of the fairness problem and the effects of heterogeneity, we consider the architecture and lexicon presented in Figure \ref{fig:generalFL}, which are typical to most FL systems. This considers that in each training round, $k$ out of $K$ total rounds, the server selects a set of participating clients, $S_k$, indexed by $n$ from the set of $C$ available clients, $S$, indexed by $N$ where $S_k \subseteq S=\{c_1,c_2,…,c_N\}$. It is assumed that each client has a private dataset, $D_n$ and that all clients are attempting to learn a model with the same structure despite potential differences in size and data distribution of the client's datasets. For each round, $k$, the server sends the global model parameters, $\theta_k$ to each of the participating clients, which perform $E$ epochs/ local rounds of training to obtain their parameters,  $\theta_{k+1}^n$. The client transmits their parameters,  $\theta_{k+1}^n$ back to the server, which, in the case of the FedAvg approach, performs weighted averaging to obtain the next round parameters, $\theta_{k+1}$.\\

When defining a system, it is typical in FL to use the terms, cross-device, and cross-silo to differentiate between data settings\cite{Nair2022}. Once a FL approach is realised, evaluation metrics are used to measure the performance of the models in the federation. Evaluation is challenging due to the private nature of the distributed data. Federated evaluation is used in this paper, whereby each client retains a proportion of its samples to construct a personal test-set for evaluation - the clients each inform the server of their individual performance. This is in contrast to centralised evaluation where the server evaluates the model using a test-set at the server (obtaining a representative dataset would involve violating client data privacy). The term, federated analytics, generalises federated evaluation for any metric\cite{Ramage2020}. Analytics, such as those presented enable server-side insights into the behaviour, performance and in this case, fairness of the clients in a federation. To achieve this, the server instructs the clients to perform certain calculations on their data, in rounds where they are selected, and the clients return the results to the server. Typically, only basic metrics are collected, limiting the explainability and transparency of existing systems. \\

FL comes in many flavours, the scope of this paper is constrained by a set of assumptions about the system architecture, training models, clients and privacy. The following assumptions are surmised from this point when referring to any aspect of a FL system and they are comparable with the majority of the literature studied:
\begin{itemize}
    \item \textbf{Centrally orchestrated} - there exists a single central entity that organises the learning, is responsible for initiating training rounds, aggregating the models and selecting clients. This is referred to as the server and orchestrator interchangeably and will be assumed as trustworthy. Implementations using distributed or blockchain control are out of scope for this paper. 
    \item \textbf{Horizontal} - horizontal FL is assumed, where the feature space is consistent across clients, as is the case in most FL applications including Google’s Gboard\cite{Hard2018} This assumption is for simplicity, enabling a focus on fairness without considering the problem of entity-alignment in vertical FL.  
    \item \textbf{Task Agnostic} - this work does not have a specific application in mind and the optimisation of the model is considered to be out of scope. 
    \item \textbf{Known Sensitive Attributes} - the set of labels corresponding to protected groups, $A$, must be known by the clients and server in order to be measured.
    \item \textbf{Blackbox Clients} - no information is available about the clients, including details of database size, dataset distribution, intended participation rate, communication capability, processing ability. However, it can be assumed that the client is capable of processing the model in question.
\end{itemize}

\section*{Related Work}
\subsection*{Existing Definitions of Fairness}
The matter of fairness is a well-established, albeit unsolved problem in centralised machine learning scenarios, where a single model is trained on a single dataset\cite{Chouldechova2018,Mehrabi2021,Pessach2022}. Given FL systems are defined by the interconnection of a number of smaller centralised systems, some traditional definitions are still relevant to federated settings, enabling comparison of some aspects of fairness between clients. These include some field wide definitions, such as \emph{'fairness through awareness"} (whereby the model should be expected to behave similarly for similar inputs) and those under the umbrella of \emph{'group fairness'} which seek to protect sensitive groups from bias by ensuring types of statistical parity across groups. Despite having some similarities, FL demands its own set of definitions of fairness to encapsulate its additional complexity.\\

Existing research offering approaches for fair-FL define fairness narrowly, as such, the more general definitions from the recent, fairness-specific, survey papers are considered as a starting point. These break fairness into a number of notions either describing symptomatic conditions that would indicate unfairness or causal mechanisms that would lead to fairness. Symptom-based definition is favourable as notions are measurable without knowledge of the system’s design and do not mandate the use of any specific techniques to achieve fairness. The definition from Shi et al. is limited as it is built on a combination of mechanisms and symptoms of unfairness which have many co-dependencies and overlaps, limiting intuitiveness\cite{Shi2023}. Rafi et al. have a purely mechanism based definition\cite{Rafi2023}. The definition offered by Vucinich et al. is symptom based, meaning it is able to encompass a wide variety of specific sources of unfairness in three easily understood and observable types\cite{Vucinich2023}. The approach is however limited, as although it considers parity in the local objective of the client’s, it fails to consider the potential unfairness if the clients fail to also solve the global objective function for the orchestrator - which is the aim of the federation, as measured by either centralised or federated evaluation. This is particularly relevant in cross-device settings where the server aims to train a model using hundreds or thousands of devices to deploy to millions. The existing definitions have the following limitations, but pave the way for a more comprehensive, measurable and symptom-driven definition:
\begin{itemize}
    \item Failure to embrace notions that enable heterogeneous systems to be fair, demanding uniformity or equality as opposed to proportionality.
    \item The lack of acknowledgment for the relevance of client contribution in quantifying fairness, building on the point above. Most approaches consider uniform client performance as fair irrespective of whether the client is 'free-riding' or behaving maliciously, for example.
    \item The use of mechanism driven notions, which suggest that the application of a particular technique, such as "minimising the inequity among clients at different points in time" from expectation fairness, is sufficient to ensure fairness without observing the systems behaviour\cite{Shi2023}.
    \item Ignorance of fairness with respect to the server. A fair federation should also incentivise the orchestrator of the learning.
    \item Lack of clear attribution of mathematical metrics to the descriptive notions.
\end{itemize} 

\subsection*{Methods to Measure Fairness}
To measure fairness, metrics are required. Extensive research to discover metrics and definitions for fairness in centralised machine learning was necessitated by vast deployment in applications that are protected by anti-discrimination legislation\cite{Hardt2016}. FL must go through the same transition in order to pose as a viable alternative with distributed data. Despite the additional complexity with FL, some of the centralised metrics to detect algorithmic bias, are still insightful in the context of individual client's performance\cite{Mehrabi2021,Pessach2022}. The most prominent metrics under the umbrella of 'group fairness' are disparate impact and demographic parity, alongside equalised odds and equal opportunity which were designed to overcome the shortcomings of the former two metrics\cite{Hardt2016}. Equalised odds is a metric that measures the difference in the true positive and false positive rates between two groups, one with and one without a binary sensitive attribute. Equal opportunity is the constrained case of equalised odds which only considers the true positive rates and is therefore less robust. For a binary sensitive attribute, $a \in A=\{a_0, a_1, ..., a_m\}$ with a classifier's binary prediction, $\hat{Y}$ and corresponding ground truth, $Y$, equalised odds is satisfied for attribute the $a$, if Equation \ref{eqn:eoddef}, holds. These metrics have been used with FL in the AgnosticFair\cite{AgnosticFair} and FCFL\cite{Cui2021} approaches. Equalised odds based measures are used in this work to evaluate sensitive group performance at the client level.

\begin{equation}\label{eqn:eoddef}
   P(\hat{Y} = 1 | a = 0, Y=y) = P(\hat{Y} = 1 | a = 1, Y=y),\;\;\;y \in \{0,1\}
\end{equation}

Uniformity over certain metrics can be indicative of fairness. Average variance is used in the well renowned FL approaches of q-FedAvg\cite{qFedAvg} and FedFV\cite{FedFV} to evaluate uniformity of performance (usually, performance in FL refers to client accuracy or efficiency\cite{Shi2023} - client $n$ is denoted as achieving performance $x_{n,k}$ in round $k$), which they attribute to fairness. A drawback of using variance based measures is the infinite range and lack of normalisation - this makes it difficult to compare approaches with different statistical characteristics. An alternative measure of uniformity is Jain's Fairness Index (JFI)\cite{Jain1998}, which is commonly used to evaluate resource allocation for computer networks using Transmission Control Protocol (TCP). In the existing adaptations of JFI for FL, the client's performance, is used as the independent variable, of which the uniformity is to be measured. The UCB-BS approach\cite{UCB-CS}, uses JFI to measure fairness and can be adapted for personalised models\cite{Divi2021}. As defined in Equation \ref{eqn:JFI}, JFI is a bounded, non-linear function of the coefficient of variation (the ratio of the population standard deviation to the population mean) where $J(x) \in (0,1]$ and for $|S_k|$ clients, its maximal value of unity corresponds to all clients performing equally and a worst case of $1/|S_k|$ occurs when the results are entirely non-uniform with respect to a variable, $x$. JFI is used in a novel way to measure uniformity in this work. A similar metric to JFI is the Gini Coefficient, $G$ - which, first proposed in 1912, is a measure of inequality, obtained from the Lorenz curve, where $G \in [0,1]$\cite{Gini1912}. Its linearity and proven track record in measuring income inequality make it an interesting metric to explore in the future, it has in some cases, been applied to machine learning\cite{Leonhardt2018}. The Gini coefficient is not used in this work as differences can be drawn between the notions of inequality and fairness\cite{Chen2023}. 

\begin{equation}\label{eqn:JFI}
   J(x) = \frac{(\sum_{i=1}^{|S_k|} x)^2}{|S_k|\sum_{i=1}^{|S_k|} x^2}
\end{equation}

Contribution-aware fairness seeks proportionality not solely parity. As clients classically only transmit their model weights to the server in FL - measuring how much any client has added to the learning process is challenging. A contribution evaluation in FL is broadly described as being a technique to measure how each client contributed to the global model - this data can be used to orchestrate a fairer system in the presence of all four flavours of heterogeneity. A number of techniques, ranging from clients self reporting (vulnerable to clients reporting inaccurate results in the knowledge that there is no mechanism for server validation) to reputation mechanisms and utility-game based schemes can be used to calculate contribution\cite{Shi2023}. Shapley values originated in game theory\cite{Shapley1951}, and are already established as a technique for XAI\cite{Das2020}. However, they can be used in FL to measure the marginal contribution of a participating client in a given round, $k$ to the global model aggregated at the end of the round, $\theta_{k+1}$. The federated Shapley value, FedSV\cite{FedShapley}, $s_{n,k}$, for client $n$, round $k$ is calculated using Equations \ref{eqn:FedSV} to \ref{eqn:Shap} (if the client is not selected to participate, such that $n \notin S_k$, it is assigned zero for that round). An auxilary dataset, $D$ is used and $\ell(\theta;D)$ denotes the loss for model parameters, $\theta$ when tested on the auxiliary test-set. The final FedSV for a client, $s_n$ is the sum of its per-round Shapley values, as shown in Equation \ref{eqn:Shap}.

\begin{equation}\label{eqn:FedSV}
   s_{n,k} = 
   \begin{cases}
       \frac{1}{|S_k|} \sum_{S_i \subseteq S_k \backslash \{n\}} [U_k(S_i \cup \{n\}) - U_k(S_i)] & \;,\;\;\;n \in S_k\\
       0 & \;,\;\;\;n \notin S_k\\
   \end{cases}
\end{equation}

\begin{equation}\label{eqn:Utility}
   U_k(Z) \coloneq u_k(\theta_{k+1}^Z)\;\;\;\textrm{where}\;\;\; \theta_{k+1}^Z = \frac{1}{|A|} \sum_{n \in A}\theta_{k+1}^n
\end{equation}

\begin{equation}\label{eqn:utility}
   u_k(\theta) = \ell(\theta^k ; D) - \ell(\theta ; D)
\end{equation}

\begin{equation}\label{eqn:Shap}
   s_n = \sum_{k=1}^K s_{n,k}
\end{equation}

A major drawback of using Shapley values is that for $n$ clients participating in a round and $K$ rounds, it has complexity of $O(K2^{|S_k|})$, therefore scaling with FedSV is not feasible. However, a wide range of viable, lower-complexity approximations for FedSV and the completed-federated Shapley value that seeks to improve on FedSV\cite{CompFedShapley}. A further concern for FL is that the federated and completed-federated Shapley values require an auxiliary central test-set which is representative of the population in order to calculate the marginal contributions in an unbiased manner. This could be obtained through donations of samples from clients or by mimicking samples using generative machine learning\cite{Jeong2018}. In many FL scenarios, having such a test-set is infeasible, however - approximations have been proposed which do not require the auxilary test-set for Shapley calculation\cite{Zheng2023,Xu2021}. As this is a research paper at limited scale, true FedSV values have been used for accuracy. The cited approximations and alternative approaches should be used as required. A fairness metric has also been proposed, combining Shapley values and Pearson's correlation coefficient, but, it is limited by its sensitivity to outliers\cite{Shi2023}.

\subsection*{Fairness Conscious Approaches to FL}

A number of existing works propose novel approaches to address aspects of the fairness problem in FL. Most commonly, they target uniformity of the accuracy achieved by the clients in classification tasks. Approaches such as AgnosticFair\cite{AgnosticFair}, Ditto\cite{Ditto}, FedFA\cite{FedFA}, FedFV\cite{FedFV}, FedMGDA+\cite{FedMGDA+}, PropFair\cite{PropFair} and q-FedAvg\cite{qFedAvg} all offer alternatives to FedAvg that seek greater uniformity. In these works, uniformity is attributed to fairness, however, uniformity of performance measures are ignorant of the extent to which the client has contributed to the model training. Other approaches addressing notions of individual fairness across clients whilst requiring proportionality of the performance to the contribution include FOCUS\cite{FOCUS}, CFFL\cite{CFFL} and the CGSV reward mechanism\cite{CGSV}. As the reward for clients participating in FL is typically the performance gain obtained from the global model, such schemes that consider the contributions incentivise participation as well as encouraging fairness. \\

Most other existing approaches focus on notions of group fairness, attempting to mitigate discrimination against protected groups in FL. FairFed\cite{FairFed}, FairFL\cite{FairFL}, FMDA-M\cite{FMDA}, FPFL\cite{FPFL}, PrivFairFL\cite{PrivFairFL} and the work using Variational AutoEncoders for FL\cite{VAEGroupFairness}, all appreciate the effect that heterogeneity can have on group fairness across the federation and use unique techniques to optimise group fairness. It is important to note, similarly to the fundamental assumptions of this paper, all of these approaches rely on the clients and server knowing the sensitive characteristics from the start of the training. In these existing attempts to mitigate individual or group unfairness mentioned above, each only address one facet of the fairness problem and are hence cannot claim to be entirely fair.\\

Finally, there exist a number of approaches that attempt to address notions of both individual and group fairness. FairFB\cite{FedFB} is based on the centralised fair learning algorithm, FairBatch - it targets group fairness and can be modified for uniformity based individual fairness as well. FedMinMax\cite{FedMinMax} tackles group fairness as a classical minimax optimisation problem and shows that this also encourages individual fairness. GIFAIR-FL\cite{GIFAIRFL} uses a dynamic reweighing strategy to encourage uniformity across clients and groups in global and personalised settings. All of these are still limited as they are uniformity based and fail to consider the levels of client contributions. To summarise, although there are a number of existing approaches for fair FL, none attempt to either quantify or solve the full spectrum of fairness concerns. 

\section*{Proposed Techniques to Quantify Fairness}

This section proposes four notions that, when used in combination, encompasses a general definition of fairness in FL. The notions proposed in this section are complete, symptom driven (in order to be measurable and independent of the system's design) and are comprised of a logical definition in direct combination with an easily calculable equation, applicable to any FL system satisfying our assumptions. The fundamental building blocks for constructing the equations, are:
\begin{itemize}
    \item \textbf{Client performance} - this should be the target of the training's objective function and is typically the accuracy on an unseen test-set. For the remainder of this work, $x_{n,k}$ denotes the clients performance and is assumed to be the accuracy achieved by client $n$ in round $k$, obtained through federated evaluation, however, this could be adapted for a range of performance metrics.
    \item \textbf{Client contribution} - the federated Shapley value up to round $k$ for each client, $s_{n}$ is calculated using Equations \ref{eqn:FedSV} to \ref{eqn:Shap}. A larger positive value indicates a more significant contribution of client $n$ to the global model.
    \item \textbf{Client reward} - denoted by $r_{n,k}$, it is the cumulative reward received by the client from the server, $n$ at the beginning of round $k$. Most commonly, the reward is a global model of higher performance than the client could train itself (however, it could also be financial). In this case, the reward is the accuracy measured on a unseen test-set, after the server has distributed the model but before local training has taken place - this is naturally cumulative as the model accuracy after a given round represents the compounded training to that point. 
\end{itemize}
The following notions define fairness in FL, each has an associated $f$ value, valid $f \in (0,1]$ with a value of 1 attributed to optimal fairness, and poses a question that enables a user to understand the implications of the notion logically:
\begin{enumerate}
    \item \textbf{Individual Fairness, $f_{j,k}$} - \textit{do all clients perform proportionately to their contribution?}\\
    As already shown, although uniformity of performance is important to fairness, if the client datasets exhibit statistical heterogeneity, clients should not be expected to achieve comparable performance. In this work, individual fairness is satisfied if all clients achieve an accuracy, directly proportionate to their Shapley contributions. The gain, $G$ is defined as the performance over the contribution and this work proposes that degree of uniformity of the gain should define individual fairness - as measured using JFI from Equation \ref{eqn:JFI} and shown in Equation \ref{eqn:F_j}. In the case of independent and identically distributed (iid) data across clients, it is predicted that client's contributions will be equivalent and the metric will reduce to the classical measure of uniformity of performance. 
    \begin{equation}\label{eqn:F_j}
        f_{j,k} = J(G)\;\;\;\textrm{where}\;\;\; G = \Bigl\{ \frac{x_{n,k}}{s_{n}} \;\;\colon\;\; n \in S_k \Bigl\}
    \end{equation}

    \item \textbf{Protected Group Fairness, $f_{g,k}$} - \textit{do subgroups of the population that exhibit sensitive attributes perform equivalently to those without?}\\
    As discussed, equalised odds is the most robust and complete measure for group fairness in centralised learning, therefore it is the best starting point for calculating group fairness in FL. This section measures group fairness via an aggregation of the proposed equalised odds difference metrics derived from Equation \ref{eqn:eoddef}. Each client, $n$ is able to identify samples corresponding to the $m$ protected groups during model validation, as stated in the paper assumptions. Therefore they can calculate the difference between the true positive and false positive rates for samples with and without each binary sensitive attribute, $a \in A = \{a_0, a_1, ... ,a_m\}$. As a result, each client calculates the equalised odds difference metric, $E_n(a)$, using Equation \ref{eqn:group fairness 1} for all sensitive attributes $a \in A$ whenever it is selected for training. Note that the value liberated by Equation \ref{eqn:group fairness 1}, $E_n(a) \in [0,1]$, with a value of unity corresponding to fair treatment of the attribute, $a$.
    \begin{equation}\label{eqn:group fairness 1}
        E_n(a) = \Big|1 - \sum_{y \in \{0,1\}} P(\hat{Y} = 1 | a = 1, Y=y) - P(\hat{Y} = 1 | a = 0, Y=y) \Big|
    \end{equation}
    The clients report the difference metrics for each sensitive attribute to the server - the server calculates the means of each of the client's values (this can be carried out at the client to reduce communication load and for more privacy but transmitting all the metrics enables finer granularity in the results from the analytics tools). Group fairness over the federation is defined as median value of the difference metric means across the clients, as in Equation \ref{eqn:F_g}. The median is selected, over other methods of average as it more resilient to outliers, representing the central tendency of the federation more effectively.
    \begin{equation}\label{eqn:F_g}
        f_{g,k} = \textrm{med} \Bigg( \biggl\{\frac{1}{|A|} \sum_{a \in A} E_{n}(a) \;\;\colon\;\; n \in S_k\biggl\} \Bigg)
    \end{equation}

    \item \textbf{Incentive Fairness, $f_{r,k}$} - \textit{are clients rewarded proportionately to their contributions and in acceptable time-frames?}\\
    Reward is important in FL, as an unfair incentive scheme may discourage participation - in the commercial realm, clients may join a competitor's federation. Incentive fairness is defined as the degree of uniformity, as measured by JFI, of the reward over contribution gains of the clients. This is logical as it enables heterogeneous systems to be fair, for example, if adversarial clients with poor data are rewarded unfavourably and clients with high quality datasets are generously rewarded but both are directly proportionate to their Shapley contribution, the system's incentive scheme will be deemed as fair. The mathematical definition is provided in Equation \ref{eqn:F_r}. A subtlety of the definition is its consideration of the temporal dimension. As both Shapley values and the reward are cumulative measures, at any round that it is measured, it considers the  client's complete experience - enabling 'catch-up' or clients to be compensated retrospectively but a graph of $f_{r,k}$ through time would still show the period in which the reward was disproportionate. This relation to time is particularly important in the case of financial rewards where the delivery may be sporadic. In the case that the reward is the global model accuracy, it is important to note that $f_{r,k}$ is not equivalent to $f_{i,k}$, as the accuracy relates to the global model 'as is' - evaluated on the local test-set prior to training compared to after local training. This is crucial as the reward should not consider the client's training ability - it represents the reward received given their contribution up to the end of the round it is measured.
    \begin{equation}\label{eqn:F_r}
        f_{r,k} = J(R)\;\;\;\textrm{where}\;\;\; R = \Bigl\{ \frac{x_{n,k}}{s_{n}} \;\;\colon\;\; n \in S_k \Bigl\}
    \end{equation}

    \item \textbf{Orchestrator Fairness, $f_{o,k}$} - \textit{does the server succeed in its role of orchestrating a learning ecosystem that maximises the objective function?}\\
    In cross-device FL in particular, the server orchestrates the learning process in order to obtain a more successful model than clients could train locally, or that the server could train due to limitations in availability of suitable data or cost. For example, when an entity such as Google wants to train for GBoard, they want to use a small proportion of a population which are available and willing to participate, to train a model that can be deployed to millions of devices. As the server has invested in the process and has facilitated the learning, this work proposes that it would be unfair if, in exchange, the server did not receive a high performing model - this motivates the introduction of orchestrator fairness. Orchestrator fairness is measured as the mean average of the normalised client performance (accuracy is already normalised if this is the metric used), $\hat{x}_{n,k}$ as is shown in Equation \ref{eqn:F_o}.
    \begin{equation}\label{eqn:F_o}
        f_{o,k} = \frac{1}{|S_k|} \sum_{n \in S_k} \hat{x}_{n,k}
    \end{equation}
\end{enumerate}

The presented notions of fairness address the limitations of previous existing definitions, are symptomatically defined and simply quantified. By collecting these metrics, a new level of explainability can be added to FL systems, at varying levels of granularity, as a large amount of data can be collected and processed at the edge - enabling client, or even sensitive attribute level detail. At the other end of the spectrum, \textbf{general fairness}, $F_T$ is presented in Equation \ref{eqn:f_T} to attribute the fairness to a single number $\in (0,1]$, it is a weighted sum of the four notions of fairness and Equation \ref{eqn:general f_T} represents the special case, in which all notions of fairness are equally weighted. Equation \ref{eqn:f_T} is used where there is premise for application specific prioritisation of specific notions of fairness.
    \begin{equation}\label{eqn:f_T}
        F_T = f_j\omega_j + f_g\omega_g + f_r\omega_r + f_o\omega_o\;\;\;\textrm{where}\;\;\; \omega_j + \omega_g + \omega_r + \omega_o = 1
    \end{equation}
        \begin{equation}\label{eqn:general f_T}
        F_T = \frac{f_j + f_g + f_r + f_o}{4}
    \end{equation}

\begin{center}
\begin{table}[h]
    \centering
    \caption{\centering Analysis of the most prominent fair approaches identified in the field of FL. The relation between each approach and the addressed notions of fairness from this work is shown. Common models are abbreviated to MLP (multi-layer perceptron), CNN (convolutional neural network), LSTM (long shirt-term memory) and SVM (support vector machine).}
    \small
    \label{table:approaches}
    \begin{tabularx}{\textwidth}{c|c|c|c|c|c}
        \hline
             \multirow{2}{*}{\textbf{Approach}} & \multicolumn{4}{|X|}{\centering \textbf{Fairness Notions}} & \textbf{Data Settings} \\
             \cline{2-5}
         & $f_i$ & $f_g$ & $f_r$ & $f_o$ & \textbf{and Models}\\
         \hline
         \smallskip\begin{minipage}[c]{55mm} \centering \textbf{FedAvg} \cite{McMahan2017}, 2017 \\ Published in PMLR\\ Contributions from Google\\ 15071 citations \\  \end{minipage} &
         \begin{minipage}[c]{6mm} \centering \textcolor{red}{$\times$} \end{minipage} &
         \begin{minipage}[c]{6mm} \centering \textcolor{red}{$\times$} \end{minipage} &
         \begin{minipage}[c]{6mm} \centering \textcolor{red}{$\times$} \end{minipage} &
         \begin{minipage}[c]{6mm} \centering \textcolor{green}{$\checkmark$} \end{minipage} &
         \begin{minipage}[c]{70mm} \centering  \textbf{MNIST} (MLP): iid and non-iid\\ \textbf{CIFAR-10} (CNN): iid \\ \textbf{Shakespeare} Dataset (LSTM): non-iid  \end{minipage}\\
         \hline
          \smallskip\begin{minipage}[c]{55mm} \centering \textbf{q-FedAvg} \cite{qFedAvg}, 2020 \\ Published in ICLR\\ Contributions from Facebook and CMU\\ 768 citations \\  \end{minipage} &
         \begin{minipage}[c]{6mm} \centering \textcolor{green}{$\checkmark$} \end{minipage} &
         \begin{minipage}[c]{6mm} \centering \textcolor{red}{$\times$} \end{minipage} &
         \begin{minipage}[c]{6mm} \centering \textcolor{red}{$\times$} \end{minipage} &
         \begin{minipage}[c]{6mm} \centering \textcolor{green}{$\checkmark$} \end{minipage} &
         \begin{minipage}[c]{70mm} \centering  \textbf{Synthetic} (Regression): iid and non-iid\\ \textbf{Vehicle} (SVM): natural non-iid \\ \textbf{Sent140} (LSTM): natural non-iid \\\textbf{Shakespeare} Dataset (LSTM): natural non-iid  \\ \textbf{Omniglot} (CNN): non-iid. \\ \end{minipage}\\
         \hline
         \smallskip\begin{minipage}[c]{55mm} \centering \textbf{Ditto} \cite{Ditto}, 2021 \\ Published in PMLR\\ Contributions from Facebook and CMU\\ 642 citations \\  \end{minipage} &
         \begin{minipage}[c]{6mm} \centering \textcolor{green}{$\checkmark$} \end{minipage} &
         \begin{minipage}[c]{6mm} \centering \textcolor{red}{$\times$} \end{minipage} &
         \begin{minipage}[c]{6mm} \centering \textcolor{green}{$\checkmark$} \end{minipage}&
         \begin{minipage}[c]{6mm} \centering \textcolor{green}{$\checkmark$} \end{minipage} &
         \begin{minipage}[c]{70mm} \centering  \textbf{Fashion-MNIST} (CNN): non-iid\\ \textbf{FEMNIST} (CNN): non-iid and natural non-iid \\ \textbf{Vehicle} (SVM): natural non-iid \\ \textbf{CelebA} (CNN): natural non-iid \\ \textbf{StackOverflow} (regression): natural non-iid \end{minipage}\\
         \hline
         \smallskip\begin{minipage}[c]{55mm} \centering \textbf{CGSV} \cite{CGSV}, 2021 \\ Published in NeurIPS\\ Contributions from the University of Singapore, SONY AI and the University of Georgia\\ 60 citations \\  \end{minipage} &
         \begin{minipage}[c]{6mm} \centering \textcolor{green}{$\checkmark$} \end{minipage}&
         \begin{minipage}[c]{6mm} \centering \textcolor{red}{$\times$} \end{minipage} &
         \begin{minipage}[c]{6mm} \centering \textcolor{green}{$\checkmark$} \end{minipage} &
         \begin{minipage}[c]{6mm} \centering \textcolor{green}{$\checkmark$} \end{minipage} &
         \begin{minipage}[c]{70mm} \centering  \textbf{MNIST} (CNN): non-iid\\ \textbf{CIFAR-10} (CNN): niid \\ \textbf{Movie Review} (CNN): non-iid  \\ \textbf{Standford Sentiment Treebank} (CNN): non-iid \end{minipage}\\
         \hline 
         \smallskip\begin{minipage}[c]{55mm} \centering \textbf{FedMinMax} \cite{FedMinMax}, 2022 \\ Published in ACM\\ Contributions from UCL, Duke University and Apple Inc.\\ 31 citations \\  \end{minipage} &
         \begin{minipage}[c]{6mm} \centering \textcolor{green}{$\checkmark$} \end{minipage} &
         \begin{minipage}[c]{6mm} \centering \textcolor{green}{$\checkmark$} \end{minipage} &
         \begin{minipage}[c]{6mm} \centering \textcolor{red}{$\times$} \end{minipage} &
         \begin{minipage}[c]{6mm} \centering \textcolor{green}{$\checkmark$} \end{minipage} &
         \begin{minipage}[c]{70mm} \centering   \textbf{Synthetic} (MLP): iid and non-iid\\ \textbf{Adult} (MLP): iid and non-iid\\ \textbf{Fashion-MNIST} (CNN): iid and non-iid \\\textbf{CIFAR-10} (ResNet-18): iid and non-iid  \\ \textbf{ACS Employment} (MLP): iid and non-iid\end{minipage}\\
    \hline
    \end{tabularx}
    \end{table}
\end{center}
\normalsize

\section*{Experimental Design}
\begin{figure}[h]
    \centering
    \includegraphics[scale=0.7]{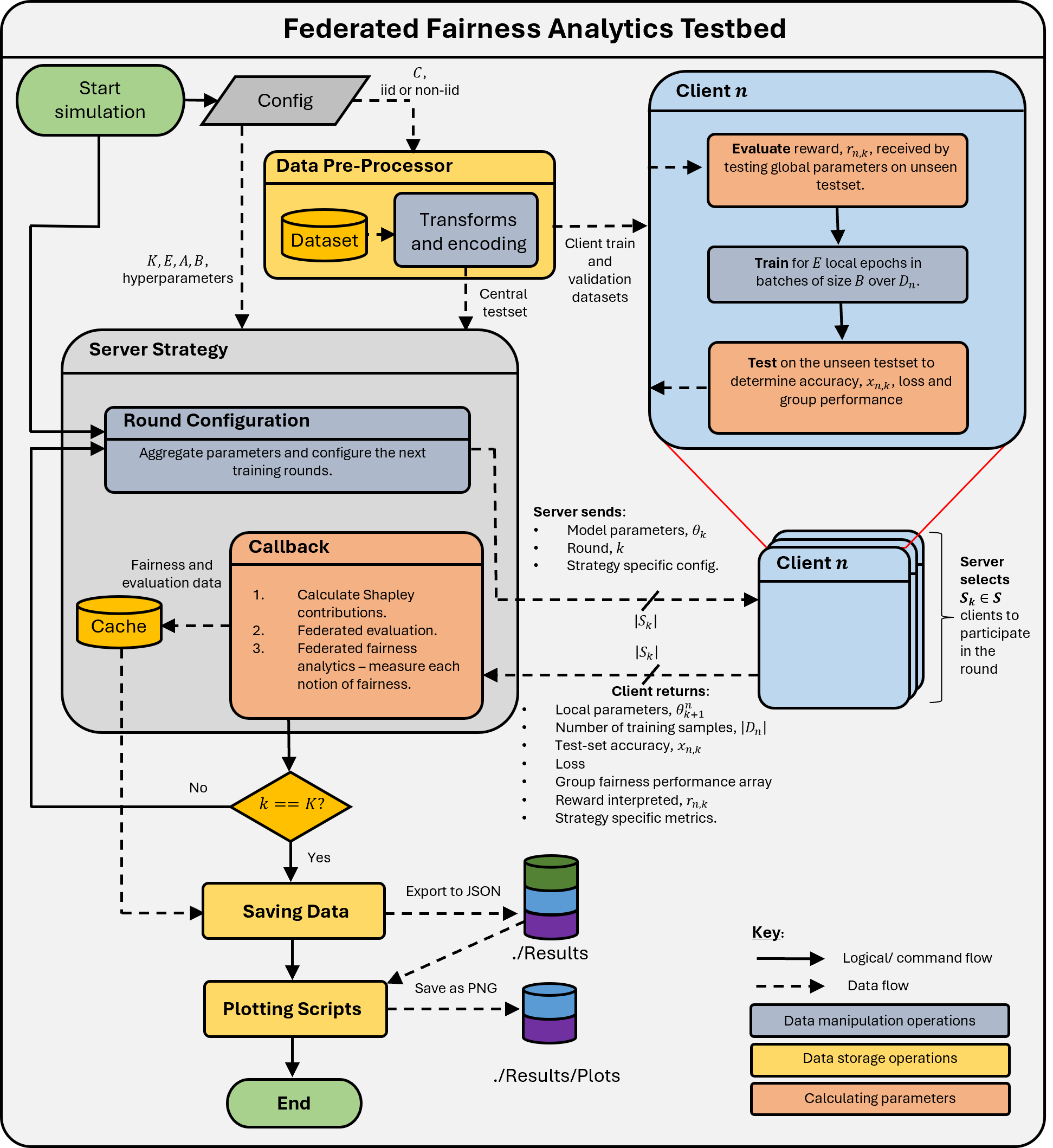}
    \caption{\centering A logical schematic of the simulation testbed.}
    \label{fig: implementation}
\end{figure}

The experimental work in this paper aims to test the metrics for fairness analytics, on a wide range of FL scenarios. Many factors could impact fairness including the model, the dataset, heterogeneity conditions, the number of clients, the client sampling rate, the values of hyperparameters or the application of privacy-enhancing techniques. This section justifies the selection of the most prominent of these for experimentation, as displayed in Table \ref{table:experiments}. Figure \ref{fig: implementation} shows the architecture of the testbed used to simulate the differing conditions and to calculate fairness during training time. 

\begin{center}
\begin{table}[h]
    \centering
    \caption{\centering Summary of the conditions for the Twenty-Four experimental conditions.}
    \small
    \label{table:experiments}
    \begin{tabularx}{\linewidth}{ccccccc}
    \hline
        \multirow{2}{*}{\textbf{Strategy}} & \multirow{2}{*}{\textbf{Dataset}} & \multirow{2}{*}{\textbf{Settings}} & \multirow{2}{*}{\textbf{$|S_k|$}} & \multirow{2}{*}{\textbf{$\mu_s$}} & \multirow{2}{*}{\textbf{$E$}} & \multirow{2}{*}{\textbf{Sensitive Attributes, $A$}} \\
        & & & & & &  \\
    \hline
    \multirow{3}{*}{\begin{minipage}[c]{28mm} \centering \textbf{FedAvg}\end{minipage}} &
    FEMNIST&
    iid, natural non-iid&
    205&
    2.43\%&
    5&
    Digits \\
    & % Blank for multirow
    CIFAR-10&
    iid, non-iid&
    10, 100&
    50\%, 5\%&
    10&
    All classes \\
    & % Blank for multirow
    NSL-KDD&
    iid, non-iid&
    100&
    5\%&
    5&
    Anomalous traffic \\
    \hline
        \multirow{3}{*}{\begin{minipage}[c]{40mm} \centering \textbf{q-FedAvg}\end{minipage}} &
    FEMNIST&
    iid, natural non-iid&
    205&
    2.43\%&
    5&
    Digits \\
    & % Blank for multirow
    CIFAR-10&
    iid, non-iid&
    10, 100&
    50\%, 5\%&
    10&
    All classes \\
    & % Blank for multirow
    NSL-KDD&
    iid, non-iid&
    100&
    5\%&
    5&
    Anomalous traffic \\
    \hline
        \multirow{3}{*}{\begin{minipage}[c]{40mm} \centering \textbf{Ditto}\end{minipage}} &
    FEMNIST&
    iid, natural non-iid&
    205&
    2.43\%&
    5&
    Digits \\
    & % Blank for multirow
    CIFAR-10&
    iid, non-iid&
    10, 100&
    50\%, 5\%&
    10&
    All classes \\
    & % Blank for multirow
    NSL-KDD&
    iid, non-iid&
    100&
    5\%&
    5&
    Anomalous traffic \\
    \hline
    \end{tabularx}
    \end{table}
\end{center}
\normalsize

\subsection*{FL Approaches} \label{subsubsec:approaches}
As discussed, a number of existing FL approaches attempt to address different aspects of fairness. The aim, is to evaluate a pair of the most promising approaches from the literature against the baseline approach, FedAvg. In order to select the most interesting benchmarks, all known approaches that primarily target fairness were qualitatively evaluated against our four notions of fairness. By analysing each publication, it was determined which of the logical definitions of fairness that the approach attempted to address (only the awareness/intent is assessed, not the success of the approach) and the key techniques used were noted. From this, the resulting frameworks in Table \ref{table:approaches} were selected as the most interesting based on their provenience (publication, academic or commercial institutions and number of citations on Google Scholar), and the range of fairness notions that they addressed. For the additional approaches - \textbf{q-FedAvg} is selected as it is commonly accepted as the first approach seeking fairness and appears extensively in the literature. \textbf{Ditto} is also selected for variety - testing on a personalisation based algorithm. CGSV and FedMinMax were discarded as they were not compatible with our assumptions - FedMinMax requires all clients to participate in every round of training and CGSV uses model sparsification (model adaptation is out of the scope of this paper and would reduce the comparability between benchmarks).

\subsection*{Data Sets and Models} \label{subsubsec:data settings}
The following labelled datasets and their corresponding classification models are selected for this work. It is important to note that optimal model design is out of scope and the lack of complete optimisation results in lower orchestrator fairness, $f_{o,k}$.
\begin{itemize}
    \item \textbf{CIFAR-10} is selected as it is one of the most commonly used datasets in machine learning experiments, it consists of 60,000 $32\times32$ pixel, 3 channel colour images of 10 classes of item. Many high performing multi-class classifiers have been designed for CIFAR-10 and this paper uses the model from the Flower tutorial\cite{Beutel2020}. The CNN consists of a 2D convolution layer with max-pooling followed by another 2D convolutional layer and three fully connected linear layers. ReLu activation functions are used.
    \item \textbf{FEMNIST}, also known as Federated-EMIST is one of the LEAF datasets\cite{LEAF}, designed for effective bench-marking of heterogeneous FL. It consists of 805,263, $28\times28$ pixel images of handwritten characters and is selected for its natural non-iid partitioning. The model trained is a convolution neural network (CNN), comprised of two 2D convolution layers each with leaky ReLu activation followed by max-pooling layer, then by a linear layer with dropout and a fully connected layer outputting to 62 bins corresponding to the 62 FEMNIST classes (lowercase letters, capital letters and digits).
    \item  \textbf{NSL-KDD} was selected in order to further vary the test data-types and to address an area of personal interest - network intelligence. The NSL-KDD dataset\cite{NSLKDD} is comprised of 185,400 internet traffic records with 42 attributes (a mix of continuous, multi-class and binary) and is used for intrusion detection. For this paper, a binary classification model was designed, consisting of four fully connected linear layers with ReLu activation.
\end{itemize}

Other datasets that were considered include FLAIR\cite{flair}, which was not used due to the computational requirements to process the high resolution images, MQTT IoT\cite{MQTT} for intrusion detection and language based datasets such as Shakespeare also from LEAF\cite{LEAF} which, as seen in Table \ref{table:approaches} is commonly used in FL due to its natural heterogeneity. Investigation on further datasets is a target for future work.

\subsection*{Introducing Heterogeneity} \label{subsubsec:heterogeneity}
By default, most datasets are provided as a single set with a train:test split provided. For CIFAR-10, it is 83\% : 17\%, 90\% : 10\% for FEMNIST and 81\% : 19\% for NSL-KDD. The datasets require splitting across the clients with both \textbf{iid} and \textbf{non-iid} partitioning in order to modulate the statistical heterogeneity. In order to make the experiments fair and repeatable, the strategy for data partitioning must be clearly defined. Depending on the dataset, two approaches are used to divide the data amongst clients:
\begin{itemize}
    \item The \textbf{FEMNIST} dataset is designed with natural partitions - each client can represents a different writer exhibiting differing classes and numbers of samples. The dataset has 3,550 unique users with individual train-validation splits already defined. To create the \textbf{non-iid} split, 205 clients are randomly selected as used in Ditto (all selection in the paper is pseudo-random, using default random seeds) and for the \textbf{iid} partition, data samples from the whole set are randomly selected and assigned to clients, such that each client has 227 samples, corresponding to the average number of samples per user, defined in LEAF\cite{LEAF}. From the remaining 3345 users, $205\times227\times10\% = 4654$ samples are randomly selected to form the auxiliary dataset for the server to calculate Shapley values, maintaining the 90\% : 10\% train-test ratio.
    \item \textbf{CIFAR-10} and \textbf{NSL-KDD} are originally datasets for centralised learning so do not have natural splits. The whole test-set is retained as the server's auxiliary dataset and the train set is divided entirely amongst users. For \textbf{iid} partitioning, samples are randomly divided amongst clients such that each client has the same statistical distribution and number of samples. For the \textbf{non-iid} split, partitioning based on the Dirichlet distribution is used, with $\alpha = 0.5$ for CIFAR-10 and $\alpha = 3$ for NSL-KDD, the effects on fairness of varying $\alpha$ is a direction for future work. A discussion into partitioning techniques can be seen in the work from Iyer\cite{Iyer2024}. Each client's dataset is then split into a 90\% training and 10\% unseen validation set. 
\end{itemize}
For comparability and to limit the computational complexity of Shapley calculations, all experiments sample 5 clients per round, with varying sample rates, $\mu_s$. The CIFAR-10 experiments with a population of 10 clients represents a cross-silo setting whilst the rest represent cross device.

\begin{figure}[h]
\centering
\begin{subfigure}[b]{\textwidth}
         \centering
         \includegraphics[width=0.78\textwidth]{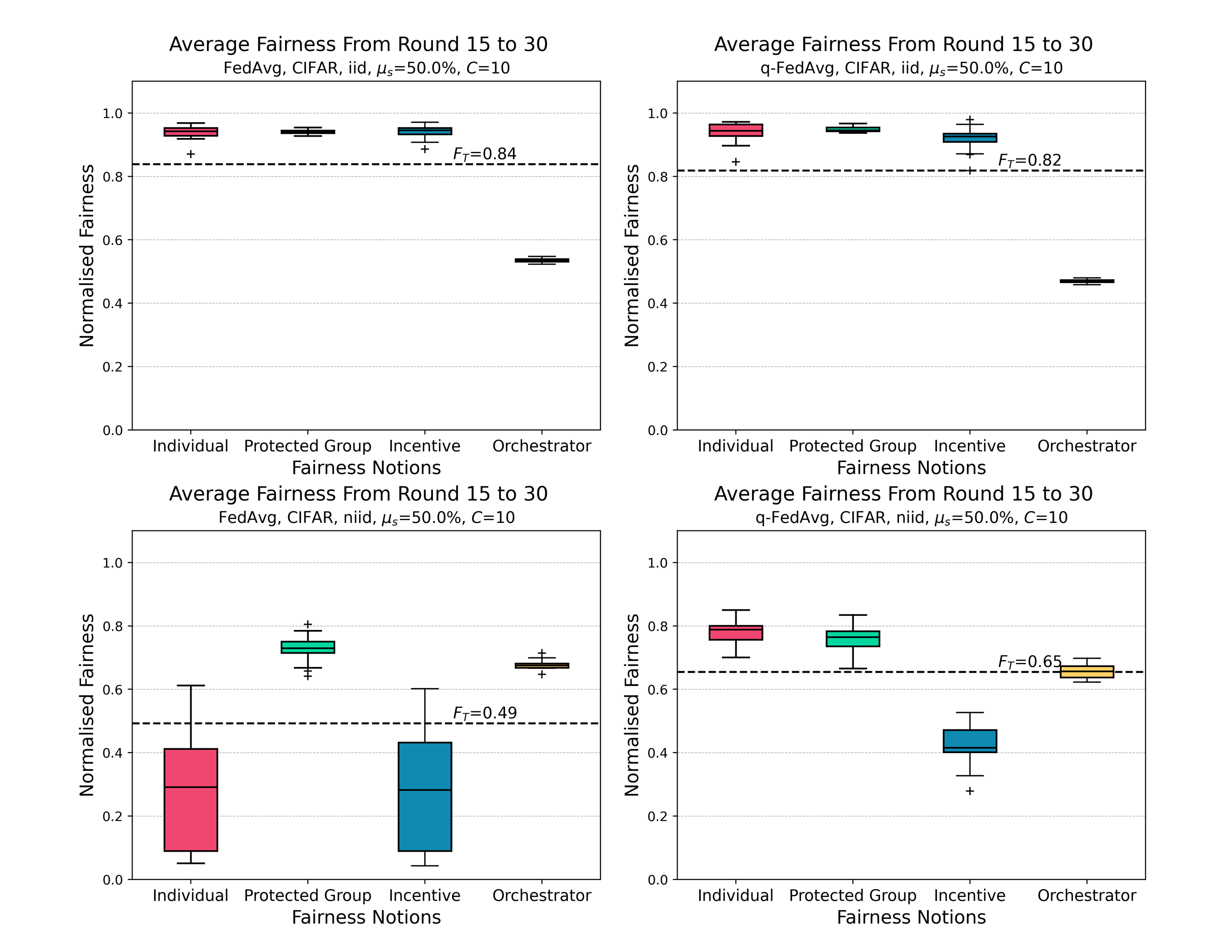}
         \caption{\centering Demonstrating the improvement in fairness for cross-silo, non-iid, CIFAR-10 training when moving from FedAvg to q-FedAvg.}
         \label{fig:results1a}
\end{subfigure}
\begin{subfigure}[b]{\textwidth}
         \centering
         \includegraphics[width=0.78\textwidth]{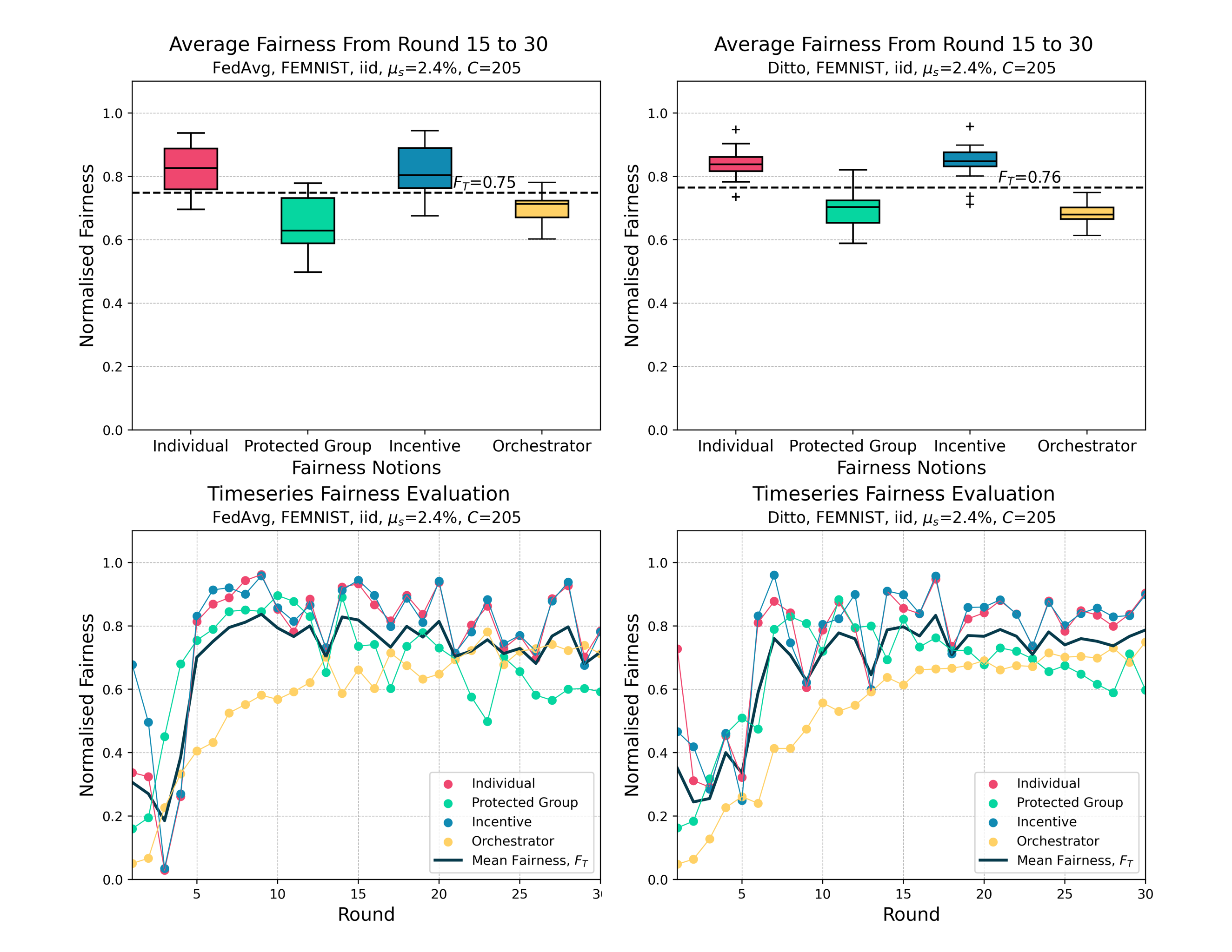}
         \caption{\centering For cross-device training of FEMNIST in the iid setting, the variation in the fairness is reduced when moving from FedAvg to Ditto.}
         \label{fig:results1b}
\end{subfigure}
\caption{Sample results at the fairness-metric level of abstraction.}
\label{fig:results1}
\end{figure}
\clearpage

\begin{figure}[h!] %options are : !htbp
    \centering
    \includegraphics[scale=0.7]{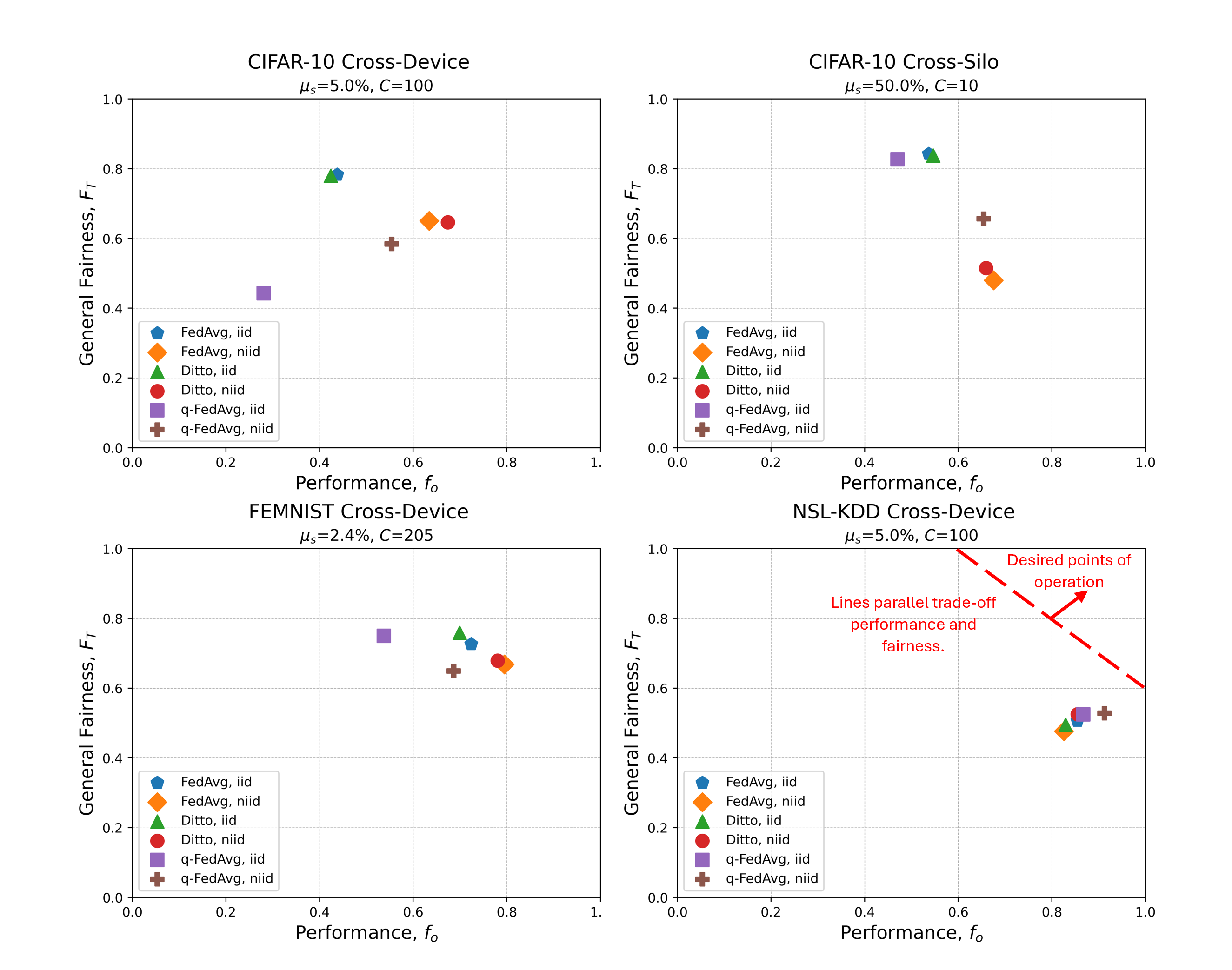}
    \caption{\centering Comparing the general fairness of different approaches against the mean-average performance.}
    \label{fig: results 2}
\end{figure}

\section*{Results and Discussion}
Federated Fairness Analytics enable the fairness of FL systems to be quantified, explained and visualised. This section displays some results from the experiments described in Table \ref{table:experiments}, in order to demonstrate important insights about the effects of the frameworks, data settings and heterogeneity on fairness, which would not be possible without this work. As the tools are designed to be a modular add-on, these results only scratch the surface of the interesting and important possible applications of these tools. Acceptable levels of fairness are subjective and application dependent, however, this work proposes that a threshold of 0.8 to constitute acceptable general fairness of a system, loosely corresponding to 80\% of users achieving fairness in the JFI based metrics. 

\subsection*{Resultant System Fairness}
The most atomic conclusions, are best visualised using the error bar and time series charts to compare experiments at the individual metric level, as in Figure \ref{fig:results1}. The results from rounds 15 to 30 are mean-averaged in order to consider how each notion of fairness settles as the model converges. It is clear that fairness is sensitive to the approach, data setting, and heterogeneity and it is anticipated that numerous other factors will affect fairness. Key, previously unobtainable insights that can be drawn from this subset of systems are as follows, demonstrating that Federated Fairness Analytics enable a unique perspective to observe systems:
\begin{itemize}
    \item The top row of Figure \ref{fig:results1a} shows that, with a cross-silo CIFAR-10 classifier in the iid setting, fairness is invariant with the change from FedAvg to q-FedAvg. This is logical as q-FedAvg seeks fairer resource allocation but each client exhibits the same distribution so q-FedAvg will tend to FedAvg.
    \item Comparing to the bottom row of Figure \ref{fig:results1a}, non-iid data distributions (statistical heterogeneity) degrade fairness performance with FedAvg as well as increasing its variance significantly. Moving to q-FedAvg, the bottom-right demonstrates a significant improvement in both general fairness and minimum fairness at any point. 
    \item Figure \ref{fig:results1b} demonstrates that for the FEMNIST dataset with iid partitioning, the variation in fairness is reduced by using Ditto instead of FedAvg. It also indicates that fairness is not static through training time, showing sporadic behaviour in early rounds, then settling - leading to the idea of fairness convergence in the training of FL systems. The full results also show that the temporal behaviour can depend on the approach, data conditions, ML task, etc. 
\end{itemize}

\subsection*{Fairness-Performance Trade-off}
The results in Figure \ref{fig:results1} are more useful for fairness tuning at later stages of deployment, whereas those in Figure \ref{fig: results 2} can be used for top-level comparison of conditions. Orchestrator fairness is analogous to performance and these figures are used to observe the trade-off between fairness and performance over the full set of experiments. By selecting implementations on the dashed lines, which show the best conditions, users can select approaches that favour fairness over performance, or vice versa, as is shown. It is particularly interesting to observe that for the FEMNIST task, Ditto directly trades performance for fairness and q-FedAvg under performs. Whereas for NSL-KDD q-FedAvg offers the best in fairness and performance. In all cases, fairness reduces with statistical heterogeneity, typically by a greater margin than a fairness-conscious approach can remedy. In the CIFAR-10 experiments, it shows that cross-device and cross-silo settings exhibit different characteristics, indicating that fairness is sensitive to the client participation rate, $\mu_s$ and dataset size.

\section*{Conclusion}

Primarily, it should be clear, that fairness is still an unsolved problem in FL. However, the objective of this work was not to resolve unfairness but to define and measure it, in order to evaluate FL systems. Building on existing literature, we developed of a set of four complete, symptom-driven notions which together encapsulate how unfairness can manifest itself in FL. Each notion is paired with a quantifiable metric; these metrics underpin Federated Fairness Analytics - the methodology to actively measure fairness at training time. The analytics enable fairness-explainable-AI by producing insights into the effects of design decisions on fairness that were previously not possible. This paper analysed the fairness performance of Twenty-Four different FL systems which vary in FL approach, ML task, data setting and heterogeneity. The results demonstrate that fairness typically drops with increases in statistical heterogeneity or decreases in client participation rate. Under our definition, fairness-conscious FL approaches, Ditto and q-FedAvg are not shown to improve fairness in iid data settings where fairness is typically high and offer only marginal improvements in fairness in some non-iid settings. This development opens avenues for a plethora of further experiments and could be a key enabler to improve fairness. Many approaches to reduce unfairness should be tested but this may include augmenting datasets with generative-AI to reduce statistical heterogeneity or utilising the measures of fairness as an optimisation objective.\\

The main limitation of Federated Fairness Analytics is the non-linearity of the metrics - this is a limitation of Jain's fairness index and may limit intuitive interpretation and comparison of results. Consideration should be given to whether linear measures such as the Gini coefficient may be more suitable than JFI and if it can be applied for all four metrics such that their numerical meaning is directly translatable. Industry hardening would also require improving scalability. The computational bottleneck in the current version arises due to calculating full Shapley values, an $O(K2^{|S_k|})$ complexity operation. This requires implementation of the Shapley approximations, which could also void the requirement for an auxiliary dataset, which, as discussed, is not obtainable in most real-world applications. To conclude, this paper summarises first works in fairness-XAI for FL and provide a number of useful benchmarks. Future experimentation may aim to collect further data on the effects of varying number of participating clients, types of heterogeneity (also modulating heterogeneity, for example through the Dirichlet $\alpha$ parameter), complexity of models and datasets, approaches and effects of privacy-enhancing techniques such as differential privacy. It would also be interesting to investigate more future-realistic FL scenarios by moving from simulation to networks of heterogeneous devices. Finally, investigating the fairness performance in systems utilising unsupervised learning, foundational models and fine-tuning is crucial to align with current and future trends in FL. 

\bibliography{references}

\section*{Supplementary Material}
\subsection*{Implementation Details and Tabular Results}

For each of the Twenty-Four experimental settings, the FL system is simulated for 30 server rounds (by training each model for an unbounded number of rounds, this was determined to achieve satisfactory convergence for this research as optimal convergence is not the objective) and after each round, the four fairness metrics, alongside their building blocks are calculated and stored. FL simulations are implemented using the Flower framework\cite{Beutel2020}, accelerated with a NVIDIA A40 GPU. Data pre-processing varies based on dataset, in order to be compatible with the training models - tensors hold float values $\in [0,1]$, therefore, all inputs, irrespective of data type must be mapped to this range. After transforming the dataset, the testset is separated and partitioned across the $C$ clients and stored as PyTorch DataLoader objects. The transform process and hyper-parameter selection differs for each dataset and are listed for reproducible results:
\begin{itemize}
    \item \textbf{CIFAR-10} requires normalisation of each of the three channels in the image. To create the non-iid data setting, $\alpha = 0.5$ is used for Dirichlet partitioning. For q-FedAvg, $q=0.2$ is used with a QFFL learning rate $\eta_q=0.1$. For the Ditto approach, hyper-parameter values of $\lambda=0.8$ and $\eta_{l}=0.01$ are selected with 10 personalised training epochs. 
    \item \textbf{FEMNIST} has a single channel that requires normalisation. The non-iid split is obtained naturally as it is a federated dataset. For q-FedAvg, $q=0.0005$ is used with a QFFL learning rate $\eta_q=0.1$. For the Ditto approach, hyper-parameter values of $\lambda=0.836$ and $\eta_{l}=0.02$ are selected with 10 personalised training epochs. 
    \item \textbf{NSL-KDD} does not have a uniform data-schema, like the images do, and requires one-hot encoding for the categorical fields and normalisation for the continuous values. To create the non-iid data setting, $\alpha = 3$ is used for Dirichlet partitioning. For q-FedAvg, $q=0.2$ is used with a QFFL learning rate $\eta_q=0.1$. For the Ditto approach, hyper-parameter values of $\lambda=0.85$ and $\eta_{l}=0.01$ are selected with 10 personalised training epochs. 
\end{itemize}

Results are stored in a custom JSON structure designed in order to handle the complex data. In this work, each experiment is repeated 3 times and the data mean-averaged to increase confidence in results.

\begin{center}
\begin{table}[h]
    \centering
    \caption{\centering Group Fairness, $F_T$ recorded for each experiment. Reporting mean average fairness from rounds 15 to 30, mean averaged over 3 repeats.}
    \small
    \label{table:results}
    \begin{tabularx}{96mm}{c|c|c|c|c|c|c}
        \hline
             \multirow{2}{*}{\textbf{Dataset}} & \multicolumn{2}{|c|}{\centering \textbf{FedAvg}} & \multicolumn{2}{|c|}{\centering \textbf{q-FedAvg}} & \multicolumn{2}{|c}{\centering \textbf{Ditto}} \\
             \cline{2-7}
         & iid & niid & iid & niid & iid & niid\\
         \hline
         & & & & & & \\
         CIFAR-10 Cross-Device & 0.78 & 0.65 & 0.44 & 0.59 & 0.78 & 0.63 \\
         & & & & & & \\
         CIFAR-10 Cross-Silo & 0.84 & 0.49 & 0.82 & 0.65 & 0.84 & 0.52 \\
         & & & & & & \\
         FEMNIST Cross-Device & 0.75 & 0.68 & 0.72 & 0.63 & 0.76 & 0.67 \\
         & & & & & & \\
         NSL-KDD Cross-Device & 0.50 & 0.49 & 0.53 & 0.53 & 0.48 & 0.53 \\
         & & & & & & \\
         \hline
    \end{tabularx} 
    \end{table}
\end{center}
\normalsize

\newpage
\subsection*{CIFAR-10 Cross-Device Results}

\begin{figure}[h!] %options are : !htbp
    \centering
    \includegraphics[scale=0.75]{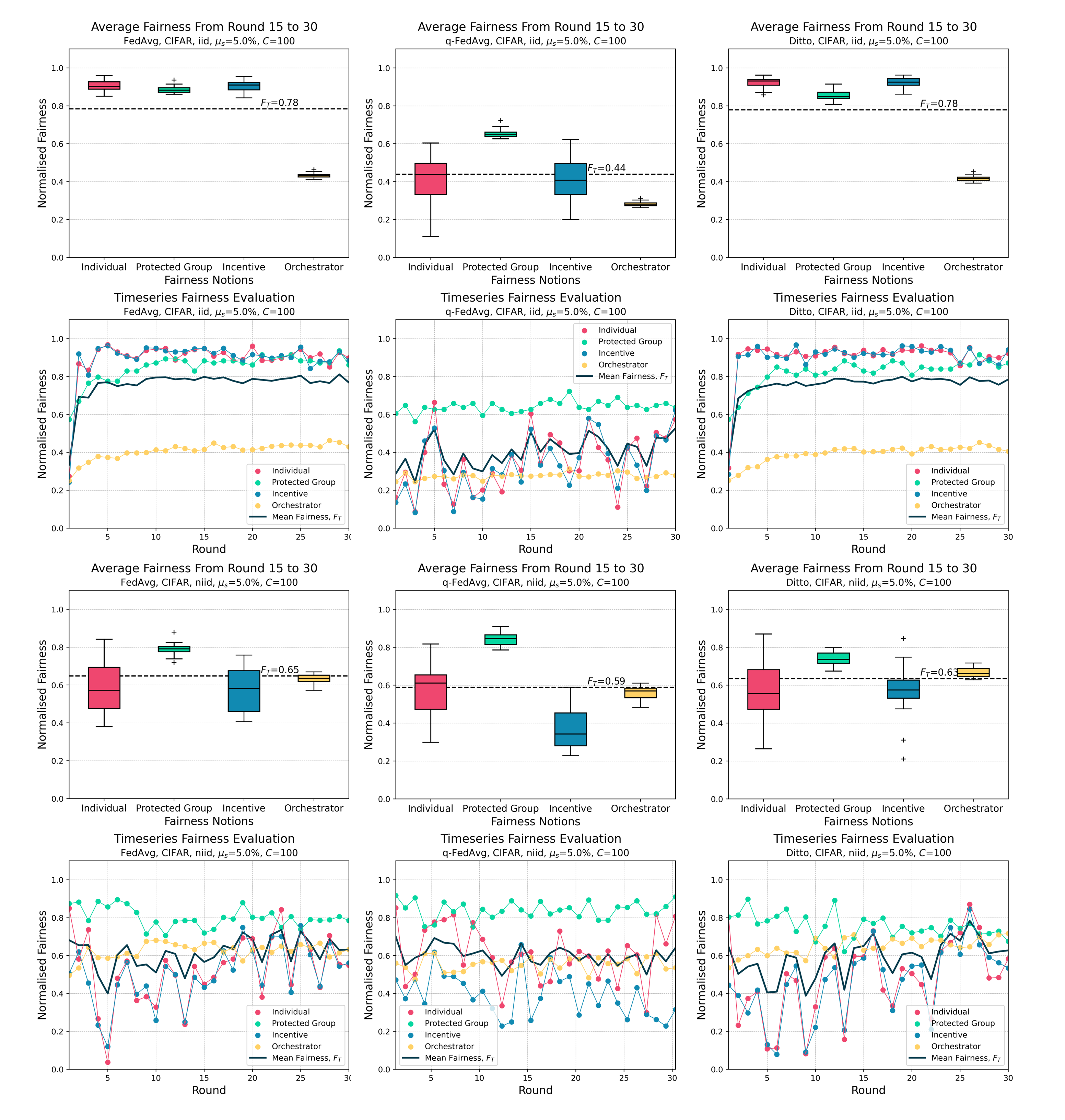}
    \caption{\centering CIFAR-10 Cross-Device Simulation Results.}
\end{figure}

\newpage
\subsection*{CIFAR-10 Cross-Silo Results}

\begin{figure}[h!] %options are : !htbp
    \centering
    \includegraphics[scale=0.75]{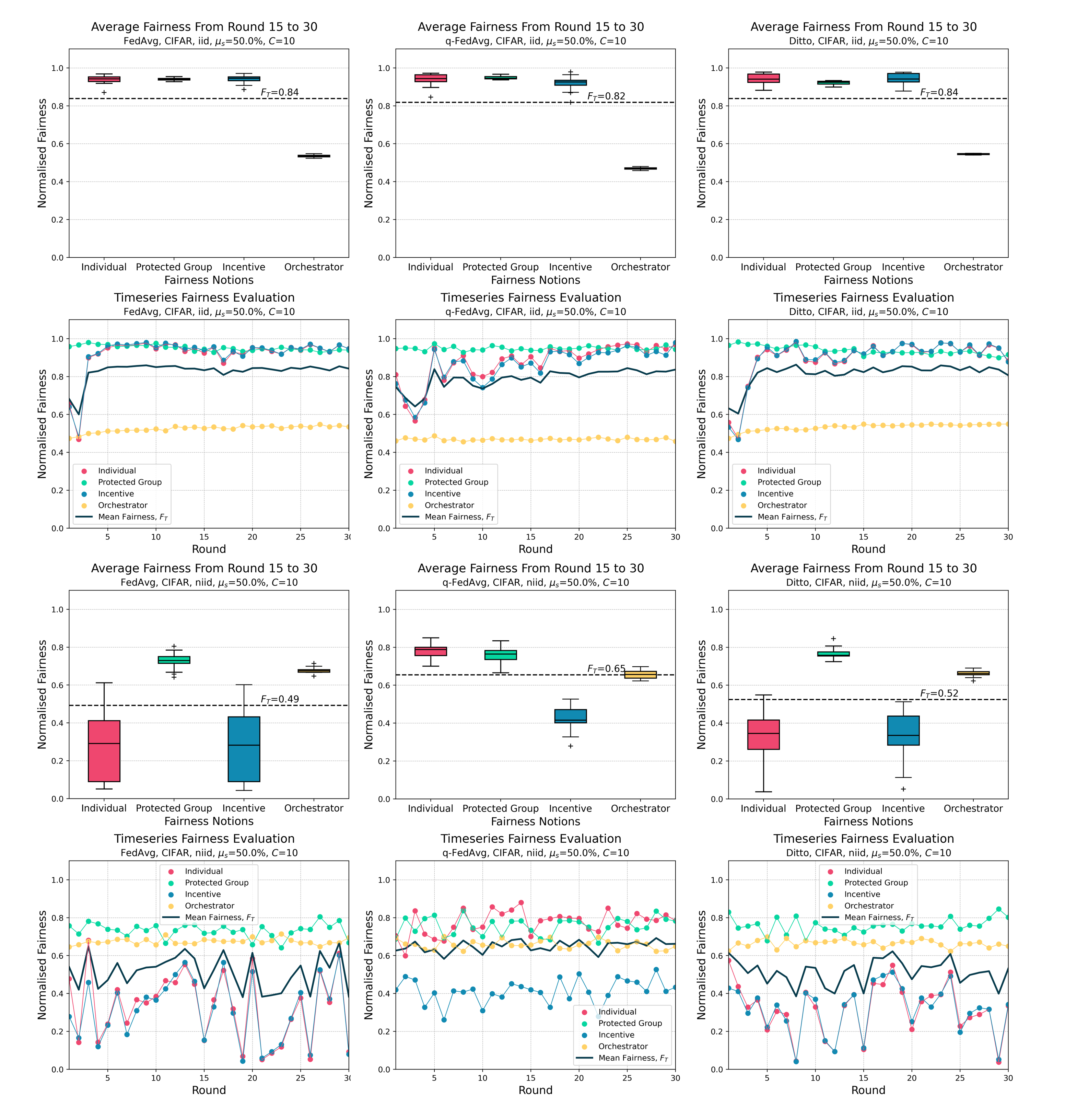}
    \caption{\centering CIFAR-10 Cross-Silo Simulation Results.}
\end{figure}

\newpage
\subsection*{FEMNIST Cross-Device Results}

\begin{figure}[h!] %options are : !htbp
    \centering
    \includegraphics[scale=0.75]{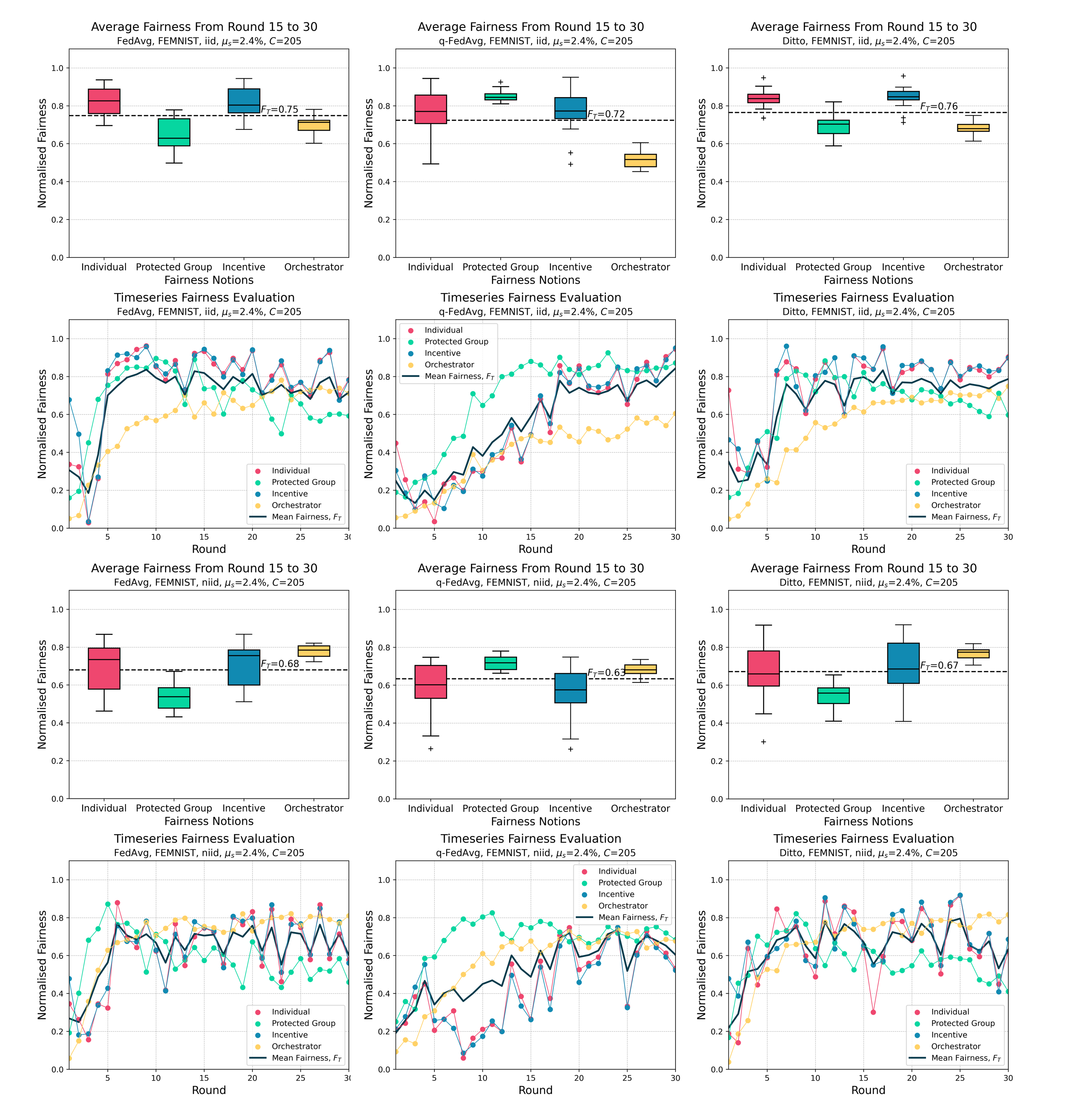}
    \caption{\centering FEMNIST Cross-Device Simulation Results.}
\end{figure}

\newpage
\subsection*{NSL-KDD Cross-Device Results}

\begin{figure}[h!] %options are : !htbp
    \centering
    \includegraphics[scale=0.75]{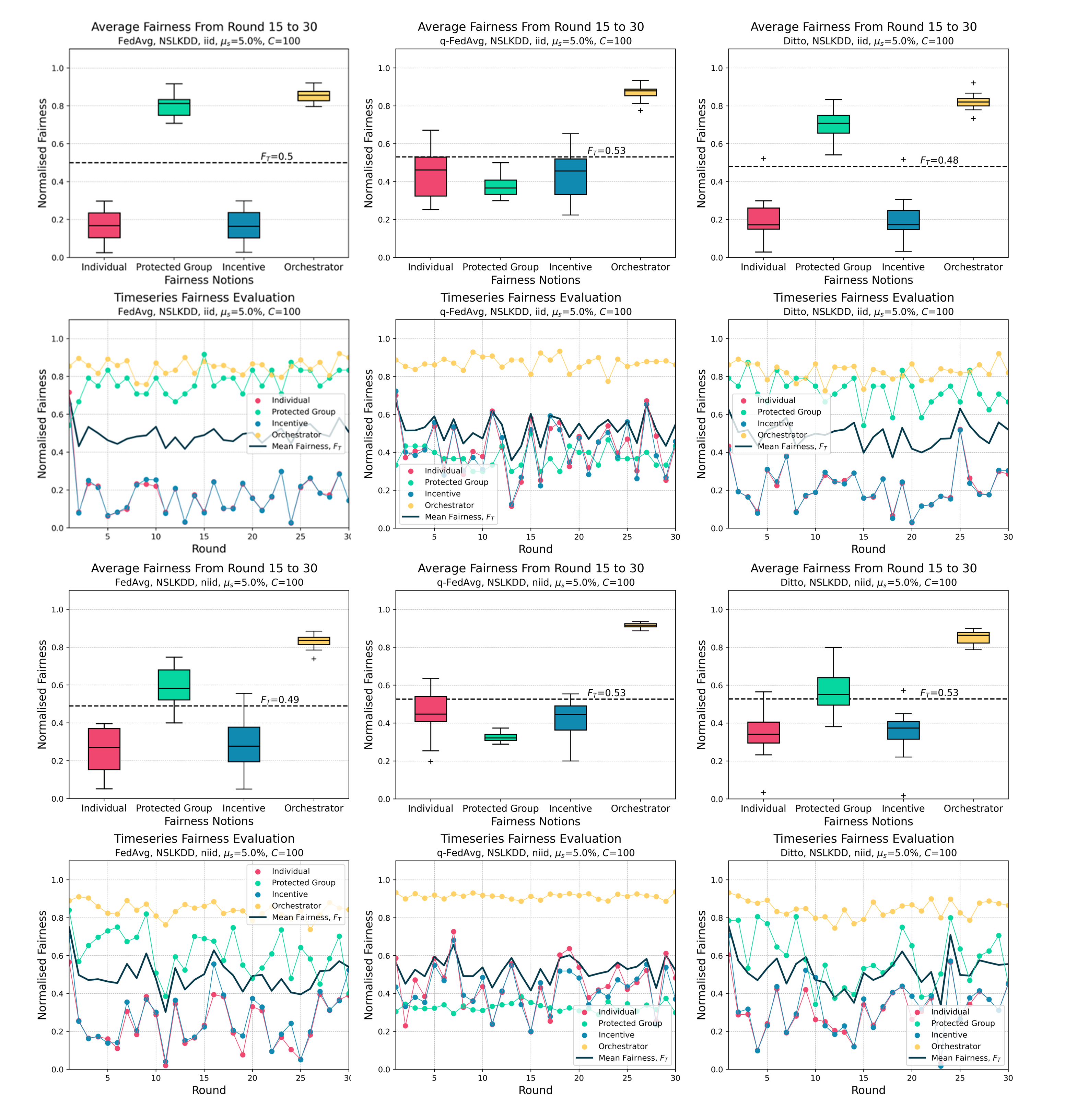}
    \caption{\centering NSL-KDD Cross-Device Simulation Results.}
\end{figure}

% \end{document}

\section*{Acknowledgements}

This work is a contribution by Project REASON, a UK
Government funded project under the Future Open Networks
Research Challenge (FONRC) sponsored by the Department
of Science Innovation and Technology (DSIT).

\section*{Author contributions statement}

The individual contributions of each author can be summarised as:
\begin{itemize}
    \item \textbf{Oscar Dilley:} Conceptualisation, Formal Analysis, Investigation, Methodology, Validation and Writing. 
    \item \textbf{Juan Marcelo Parra-Ullauri:} Conceptualisation, Validation, Review and Supervision.
    \item \textbf{Rasheed Hussain:} Funding acquisition, Conceptualisation, Validation, Review and Supervision.
    \item \textbf{Dimitra Simeonidou:} Funding acquisition, Review and Supervision.
\end{itemize}

\section*{Data Availability Statement}

Our machine learning models are trained and tested using the following open-source datasets which can be accessed via Hugging Face as follows:
\begin{itemize}
    \item uoft-cs/cifar10: \url{https://huggingface.co/datasets/uoft-cs/cifar10}
    \item flwrlabs/femnist: \url{https://huggingface.co/datasets/flwrlabs/femnist}
    \item Mireu-Lab/NSL-KDD: \url{https://huggingface.co/datasets/Mireu-Lab/NSL-KDD}
\end{itemize}

Raw data collected from our experiments, in JSON format, an archive of plots, plotting scripts and per-experiment hyper-parameter selection are made available publically on GitHub: \url{https://github.com/oscardilley/federated-fairness}.

\section*{Additional information}

The authors declare no competing interests.

\end{document}